\documentclass[10pt,twocolumn,letterpaper]{article}

\usepackage{iccv}
\usepackage{times}
\usepackage{epsfig}
\usepackage{graphicx}
\usepackage{amsmath}
\usepackage{amssymb}
\usepackage{caption}
\usepackage{subcaption}
\usepackage{tabularx}

\usepackage[pagebackref=true,breaklinks=true,bookmarks=false]{hyperref}
\DeclareUnicodeCharacter{0301}{\'{e}}

\usepackage[breaklinks=true,bookmarks=false]{hyperref}

\iccvfinalcopy 

\def\iccvPaperID{****} 

\ificcvfinal\fi

\begin{document}

\title{Unified Style Transfer}

\author{Guanjie Huang,
Hongjian He,
Xiang Li,
Xingchen Li,
Ziang Liu\\
Australian National University\\
{\tt\small \{first name.last name\}@anu.edu.au}
}

\maketitle
\ificcvfinal\thispagestyle{empty}\fi

\begin{abstract}
     Currently, it is hard to compare and evaluate different style transfer algorithms due to chaotic definitions of style and the absence of agreed objective validation methods in the study of style transfer. In this paper, a novel approach, the Unified Style Transfer (UST) model, is proposed. With the introduction of a generative model for internal style representation, UST can transfer images in two approaches, i.e., Domain-based and Image-based, simultaneously. At the same time, a new philosophy based on the human-sense of art and style distributions for evaluating the transfer model is presented and demonstrated, called Statistical Style Analysis. It provides a new path to validate style transfer models' feasibility by validating the general consistency between internal style representation and art facts. Besides,  the translation-invariance of AdaIN features is also discussed.
\end{abstract}

\section{Introduction}

Image style transfer can be defined as modifying the style of given images in the desired manner. 
It can be widely applied in computer vision tasks. For example, style transfer can be applied to real-time
image processing \cite{Gltrk2016}, transferring artworks into realistic photos \cite{johnson2016perceptual},
transferring between different artists \cite{bhalley2019artist}, style transfer in super-resolution \cite{wang2020collaborative}, style
transfer as assistance to classification problem\cite{liu2020unity}, \etc.


However, the term style transfer itself is not well-defined due to the subjective understanding of ``style''.
Conventionally, the ``style'' of an image is in contrast to its ``content''. To some extent, researchers
in style transfer agree on the definition of ``style'' of an image, \ie, the general
appearance and texture of an image excluding its semantic information (content). As style can be easily checked
visually, ablation study in style transfer is usually used as a validation method. This idea naturally
brings us the question: whether the style and content of an image can be separated to modify the style part in the desired manner and keep the content part?

Thanks to the insight from Gatys' work \cite{gatys2015neural}, i.e., Neural Style Transfer (NST), shown that Gram Matrices can separate the style and content of images in a pre-trained neural network through extensive experiments.
However, it is not enough to show that the style definition in NST is exactly the same as the 
agreed understanding of style from human beings. Huang \etal~\cite{huang2017arbitrary} proposes an alternative way of
style embedding: instead of using second-order statistics information (Gram Matrices), the task of 
style transfer is already visually satisfactory with only mean and variances in each layer of 
pre-trained neural network. There are many following research works trying to increase region-specific
statitical style transfer \cite{xu2018learning,sheng2018avatarnet}, enhance trainability of the whole model rather than using fixed pre-trained encoder \cite{huang2018multimodal},
improve speed \cite{johnson2016perceptual,ulyanov2016texture}, \etc. However, the fundamental question of whether these artificially defined styles through the neural network
have a bijective correspondence to the definition of style in common sense is still unanswered.

Another observation is that there are two transfer lines: either with two input images or one image. In the first case, we call it Image-based style transfer. It is the most well-known style transfer regime which requires both a target image with the desired content information and a style image with the desired style information.
The style transfer is implemented by transferring the style in the style image to the
target image. In the second case, the input image is transformed into some pre-defined fixed style.
We call this Domain-based style transfer. Both
lines of research demonstrate the high correlation between these two types of style transfer.
However, the exact connection between these two style transfer regimes is not well-investigated and remains
unclear at the time of writing.

In this paper, we introduce the Unified Style Transfer model (UST), which can conduct both style transfer 
regimes at the same time. To the best of our knowledge, it is the first formal justification on
the connection between the two style transfer communities. This is achieved by introducing
a simple hypothesis over the distribution of styles under the artificial definition and the human vision definition. 
Specifically, we assign a learned distribution for each Domain-based style transfer task in an arbitrary Image-based style transfer model.
As a result, both style transfer regimes are unified
because the learned distribution is generative, \ie, we can draw a sample from the distribution and send it
to the Image-based style transfer model. Note that our proposed model is a ``plug-in model'' for an
external Image-based style transfer model. Thus, our proposed model can be highly applicable
to a range of other Image-based style transfer models, as long as the external model's internal style code can be \textit{representable} by a vector..

On the other hand, thanks to the introduced distribution in UST, we propose a novel model-validation method,
called Statistical Style Analysis, as an enrichment to visual check used in most ablation study in the style transfer community. 
Instead of visually checking for some style transfer examples, we demonstrate that one could 
numerically check the statistical distance between two domains of style in order to justify whether 
these domains are similar. It also has the interesting property in the diversification of style transfer, 
since most distribution has well-understood parametrisation towards variance.
This kind of idea of formulating distance metric can be also found in \cite{kotovenko2019content,svoboda2020twostage}. But the difference between 
our work and \cite{kotovenko2019content,svoboda2020twostage} is that we focus on statistical distance and we formally justify the validation method, 
while metric learning is used as an intermediate learning step in \cite{svoboda2020twostage}.

We also prove the translation-invariant property of AdaIN style code. Specifically, AdaIN style is unchanged
under swapping of homogeneous regions, \ie, regions that contain relatively different style to the neighborhood.
This shows why the research work of introducing a region mask in \cite{xu2018learning} can be considered meaningful, as
well as the research work of style decorator in \cite{sheng2018avatarnet}.

\section{Related Work}
\label{sec:related_work}
Style transfer can be roughly divided into two types according to the transfer regime: either Image-based or Domain-based. These two communities are not growing independently but in an iterative manner.

\subsection{Image-based style transfer}
The Image-based method is the most investigated area in image style transfer, partly due to the simplification of defining style by endowing the style wanted into an image.

It is proposed that the low-level texture information of an image is stored
in the high order statistics of image pixels \cite{julesz1962visual}. This idea is further
generalised and experimented with in \cite{gatys2015neural}.
By matching the second-order statistics (represented by Gram Matrices) in a pre-trained neural network (like VGG) of the target image to the style image, image style transfer can be 
achieved visually impressively through optimisation (Neural Style Transfer, or NST). The key insight from Gatys' work is that the style information is 
stored in its second-order statistics of the pre-trained neural network up to visual satisfaction.

Based on Gatys' work, there are several improvements in terms of speed. According to works of \cite{johnson2016perceptual,ulyanov2016texture}, it is 
shown that a feed-forward neural network can replace the recursive optimisation process without loss of performance that significantly improves the speed.

Another improvement is that instead of matching the whole second-order statistics, \cite{huang2017arbitrary} points out
that it is enough to only match mean and variances. This significantly reduces dimensions of style representation and improves speed up to zero-shot. Additionally, it is proposed that
the optimisation problem of Gram Matrices can be replaced by Whitening and colouring Transformations (WCT) problem from statistics \cite{li2017universal}. According to this work, a style decorator is used to taking image region factor into account \cite{sheng2018avatarnet}.
It is proposed
that by training a GAN model on region mask, arbitrary style transfer can be achieved by using the region mask as a moderator to AdaIN \cite{xu2018learning}.

\subsection{Domain-based style transfer}
Domain-based style transfer is a classic area compared to Image-based method. It is also known as
conditional image creation or image translation. The task is to modify the style of image into
some fixed style/domain desired.

The histogram equalisation technique in image processing can be viewed as an example of
modifying image style to the ``more contrast'' one, which is usually what we want, as an image enhancement technique.
Similarly, image denoising can also be viewed as changing an image into a ``more smooth and coherent'' version. Although these examples
of changing an image's low-level statistical features are not usually considered as normal style transfer task, they are important in motivating the research area of style transfer and can be considered as generalised image style transfer or generalised image style modification.

It is proposed that using conditional instance normalisation can achieve multiple styles transfer \cite{dumoulin2016learned}. It is also possible to learn intermediate layers for each style, together with encoder and decoder, called style bank
to perform Domain-based style transfer \cite{chen2017stylebank}.

On the other hand, the application of GAN model is extensively explored in the field of Domain-based style transfer. 
The conditional GAN is used to handle multi-style/multi-domain style transfer \cite{isola2017image}
for single image input. The cycle-consistent loss is used in GAN model for unpaired data with
appealing result \cite{zhu2020unpaired}. Following this work, there is an incredible research effort
in designing the cycle-consistent loss in the case of unsupervised learning for GAN model \cite{zhu2017unpaired,kim2017learning,yi2018dualgan,liu2017unsupervised}. 
The GAN architecture is also utilised in image translation model such as image recovery \cite{yang2018towards} and cartoonization \cite{royer2018xgan}.
\section{Method}
\label{sec:approach}
In this section, we focus on the proposed UST model and the novel evaluation/validation method able to be used in the style transfer model. The section is structured in the following way: first, formally introduce the UST model; then discuss the hypotheses and limitations of our model;
next, prove the translation-invariant property of AdaIN style feature; besides, introduce the novel validation method called Statistical Style Analysis; finally, discuss the statistical distance we choose in this paper.

In this chapter, there will be several similar commonly used words, in order to avoid confusion, explain in advance as follows: AdaIN features, in short term, is the means and variances of each channel of each layer of the input image in encoder (like VGG)\cite{huang2017arbitrary}. Once it is vectorised, \ie ordered in a specific way, the style of the input image, called AdaIN style, is
represented by the corresponding vector, called AdaIN vector.

\subsection{Unified Style Transfer(UST) model}
\begin{figure*}[ht!]
  \centering
  \includegraphics[width=0.8\textwidth]{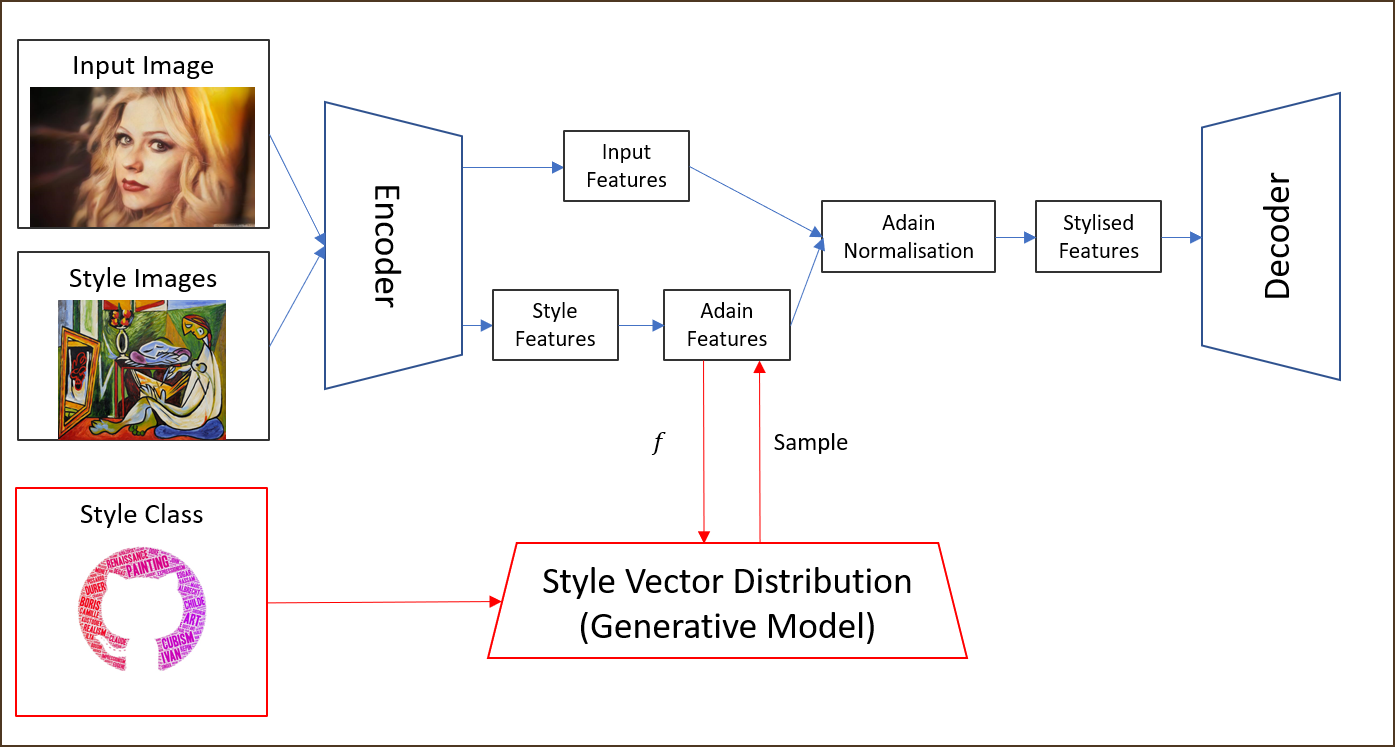}
  \caption{Overview of our approach. The UST model is the plug-in system of red blocks offering generative behaviour to an external Image-based style transfer model of blue and black blocks. This is a standard architecture for an Image-based style transfer model with \textit{representable} style. f denotes the function of formulating distribution; sample denotes drawing a sample from distribution, Style Class can be seen as any combination of style images. In our work, we use the style class provided by WikiArt, which contains groups divided by artist, style and genre.}
  \label{fig:system}
\end{figure*}

The system diagram of UST is shown in Fig.\ref{fig:system}. Note that the UST is a plug-in model (coloured by red blocks) added to
an external Image-based style transfer model.
To use the UST as a plug-in model, we require that the external model has an internal style representation
\textit{representable} by a style vector in $\mathbb{R}^n$, in which it is most of the cases.
In this paper, we choose AdaIN as the style vector for demonstration.

To achieve Domain-based style transfer in an external Image-based style transfer model, 
the style distribution of each domain is formulated over the \textit{representable} style vector space.
In this paper, we leverage a generative Gaussian model to approximate each style domain's true style distribution, also called the style label/style class. 
Domain-based style transfer is then achieved by drawing samples from the generative model.

\subsection{Fundamental hypotheses of UST model}
According to the insight of \cite{huang2017arbitrary}, there is an embedding from the space of image style, as understood visually by human beings, to the vector space of
AdaIN features. In other words, two images have a similar style visually if and only if their AdaIN features are close in the vector space.
However, \cite{huang2017arbitrary} only evaluate the reverse direction by doing a case study on a large number of style transfer experiments via visual checks.
According to this insight, we can assume that there is some uncertainty or distribution over the actual style in the sense of human vision. Therefore we inferring a distribution over the AdaIN style
vector space with the following hypothesis:
\begin{equation}
	\mbox{visually similar styles}\\
	\iff \mbox{distance close style distributions}
    \label{equation:1}
\end{equation}

Note that this hypothesis is not limited to AdaIN features. Hence this research work can be generalised easily to other style representation, as long as
the representation is \textit{representable} by a vector, and the style vector space has the property of identity of indiscernibles from the human vision perspective.

Another hypothesis we make is that the underlying distribution in the AdaIN style representation space is Gaussian. We make such a
choice because of its robustness and its simple mathematical form. However, the accurate distribution is not limited to the
specific type of distribution, for which we leave the future work. 

In order to justify the hypothesis in equation.\ref{equation:1}, we need to define the statistical distance between two distributions.
We call the process of computing and comparing the style distribution
as \textbf{Statistical Style Analysis}.

\subsection{Translation-invariance of AdaIN features}

We discover some interesting properties of AdaIN features in this paper. These properties can be proved by proof of the AdaIN vector's property, as the vectors is a specific order of the features. 

First, we find that the AdaIN features are translation-invariant. More specifically,
if we interchange two ``locally independent'' or homogeneous region of the same size in an image, 
its AdaIN vector is not changed. The term ``relatively independent'' means the convolution response
of the region's boundary does not change after swapping to another region.

We prove by induction over the number of layers. There is no channel at image input (layer 0) in the base case, so it is vacuously true.
Suppose we implement one convolution over the image, the interchange of input causes the corresponding regions interchanged at the output, 
by locally independent assumption.
Now observe that the functions mean and variance is symmetric
over permutation, especially interchanging. Thus we conclude that such an interchange does not modify the means and variances of the next layer. 
Also, any element-wise function is symmetric
over permutation. Hence the activation layer does not modify the following layer means and variances. This means
that the AdaIN style vector is unchanged under an interchanging homogeneous region.
The proof is finished by using induction.

Using a similar proof, we can also see that the style described by Gram Matrix from \cite{gatys2015neural} has a similar translation-invariant property. The meaning of ``local independence'' region
is a group of ``similar texture'' region that does not change the texture when cut. However, for Gram Matrix, it is more likely that a random region does not satisfy ``local independence'' or homogeneity because channel response's covariances may depend more on specific channels.

Another attractive property we find on the AdaIN vector is that it fails to describe ``high-level style'' in an image. We will explain that in detail in the result section.

\subsection{Statistical Style Analysis}


To justify the proposed UST model, we need to figure out the applicability of the two proposed hypotheses,
\ie, could the actual style distribution be equivalent to the style distribution defined on the \textit{representable} style vector space? And whether the Gaussian model is good enough for modelling style vector distribution for some style domains?

In order to justify the first hypothesis, we could match the statistical distance to what we understand towards the styles, as human beings. 
In other words, we assert the equation.\ref{equation:1} in a direct way:

\begin{enumerate}
\item Collect what is understood from the historical fact towards the style labels/domains;
\item Find some appropriate way of computing the statistical distances between distributions (Gaussian in this project);
\item Compare and argue whether they match, therefore justifying the applicable scope of our proposed model.
\end{enumerate}

This novel validation method has the following interesting properties over only direct visual check on images:

\begin{enumerate}
	\item The number of assertions is significantly reduced: instead of visually checking on images (infinitely many), we only need to verify each style labels (with underlying distribution), which is finitely many, thanks to the introduction of distribution over \textit{representable} style space;
	\item As a result of 1, it is possible to assert equation.\ref{equation:1} in both direction (the forward part is evaluated by picking similar styles historically and checking their statistical distance) rather than in the usual method that is only able to do the converse assertion;
\item This validation/evaluation method establish a connection between AdaIN properties and AdaIN distribution properties. More specifically, if there are some difficulties in showing some property of AdaIN style, we can instead try on its distribution to see whether the problem is easier to solve or vice versa.
\end{enumerate}

Because of the high dimension of AdaIN style vector, we provide the following update method to compute mean and covariance of Gaussian.

For covariance matrix $C$:
\begin{equation}
C_{new}=\frac{1}{m+n-1}[(n-1)C_{old}+E_{new}E_{new}^T]
\end{equation}
where $E_{new}$ is the batch matrix of vectors to be updated to $C$, m is the number of style vectors in
$E_{new}$, n is the number of all previous vectors.

The update of mean is similar:
\begin{equation}
\mu_{new}=\frac{1}{m+n}[n\mu_{old}+m\times \sum(E_{new})]
\end{equation}
where sum is taken over the number of vectors.

Through this update method, we avoid the calculation of large size matrix and save the space of system memory.

\subsection{Statistical distance}
We choose the Wasserstein metric as our distance calculation method used in Statistical Style Analysis. The reason is that it is symmetric, not sensitive to subspace difference, with the identity of indiscernibles hold, and
has closed-form formula for multivariate Gaussian distribution.
Due to space limitations, we summarise the disadvantages of some other popular statistical distance as in Table \ref{table:stat-dist} and details of comparison of different calculation method are in the Supplementary Materials.

\begin{table}
\begin{center}
\resizebox{0.44\textwidth}{!}{
 \begin{tabular}{|lp{3cm}|p{3cm}|}
 \hline
	 Statistical distance & Issue(s)\\ [0.5ex]
 \hline
 Kullback–Leibler divergence & Asymmetric \\
 Total variation distance & Almost 0 everywhere since its strong dependence on overlapping\\
 Bhattacharyya distance & Not well-defined on degenerate Gaussian; similar to total variation distance\\
 \hline
\end{tabular}}
\end{center}
	\caption{\label{table:stat-dist} Table of disadvantages of three popular statistical distances compare to the Wasserstein metric we choose.}
\end{table}

The defintion of Wasserstein metric of two probability measures $X,Y$ is as follow:

\begin{equation}
W_2(X,Y) = (\inf_\gamma\int_{V\times V} d(x,y)^2 d\gamma(x,y))^{\frac{1}{2}}
\end{equation}

For Gaussian distribution, it has the following closed-form formula:

\begin{equation}
W_2(X,Y)= ||\mu_x-\mu_y||^2_2 + tr(\Sigma_x+\Sigma_y-2(\Sigma_x\Sigma_y)^{\frac{1}{2}})
\end{equation}

\section{Experimental setup and Results}
\label{sec:Experimental setup and Results}

\subsection{Experiments}
\subsubsection{Experiment purpose}
From the methods mentioned above, our experiments mainly focus on two hypotheses: whether the visually similar styles if and only if they have close style distributions; whether the Gaussian distribution model is good enough to represent each historical class of styles such as Realism, Impressionism, \etc. To address it more specifically and easy to understand, the first hypothesis's reverse direction can be proved by visual checks in ablation study used as a normal evaluation method in other research works. We also can find evidence of whether the statistical distances of style distributions are consistent with the human understanding of art style relations to prove the forward direction. In our experiments, we leverage the historical facts and mainly focus on the forward direction as the reverse part of AdaIN features have already been proved in \cite{huang2017arbitrary}. The first hypothesis helps us validate the distributions' advantages and limitations. From a general point of view, the human's classification of paintings is not limited to the low-level style, \ie description of colours, lines, and structures of the painting. It is also influenced by geo-location, time, and culture. Therefore, statistically evaluating the result of experiments and comparing consistency with the historical style relations could give us strong feedback on distributions' performance. For the second hypothesis, experimental results can help us measure the multivariate Gaussian model's performance, which provides us with a direct insight into model suitability.

\subsubsection{Experiment methods}
To verify these two hypotheses, our experiments are divided into two sections. 
The first part of the experiment is to test the Domain-based style transfer. We set the input as a specified style label and a content image and evaluate the style transfer performance with original label images via drawing samples from the generative Gaussian model derived for each domain. The second part is to evaluate distances between style labels and determine if they obey the historical fact. To be more specific, we use the Wasserstein metric\cite{OLKIN1982257} to calculate the distance between distributions, representing different style labels. Then, we try to infer the style relationships according to the distance and compare them with the historical facts.

\subsubsection{Experiment process}
We use WikiArt dataset\cite{DBLP:journals/corr/SalehE15} as the main style image resource for experiments. The existing style label gives knowledge of the human classification of paint style. To process experiments, the AdaIN distributions for style classes are needed first. At the encoding stage, images are cropped to 512$\times$512 pixels. The encoder here is the pre-trained layers of VGG-19\cite{simonyan2014very}. We set the style vector with the length of 1920 for each image as the encoder's output. According to this, the update method calculates the mean and covariance for each style class to build the style distributions. For the first part of the experiment, we input the content image and the specified class label, then sampled AdaIn style from the corresponding distributions' mean and covariance. Finally, we can get the stylized image through the decoder.

The decoder trained by ourselves used a similar method with \cite{huang2017arbitrary} work with a bit of difference. Instead of introducing more images for training, we only use the images from the WikiArt and expect to gain better performance in decoding artworks. We only input one image for each training, which means the input image provides both content information and style information. We expect the decoder to rebuild the same image from the content feature and style feature. 

To continue the second part of our experiment, Wasserstein Metric is used to evaluate different labels' distance. Once get the distance between classes, we can leverage the historical proofs to validate whether the similar pairs are reasonable. This can be used to prove the hypothesis of the forward direction of equation.\ref{equation:1}, which if two styles have closed distributions, they should be visually similar.

\subsection{Style transfer}
\begin{figure}[htbp!]
  \centering
  \includegraphics[width=0.45\textwidth]{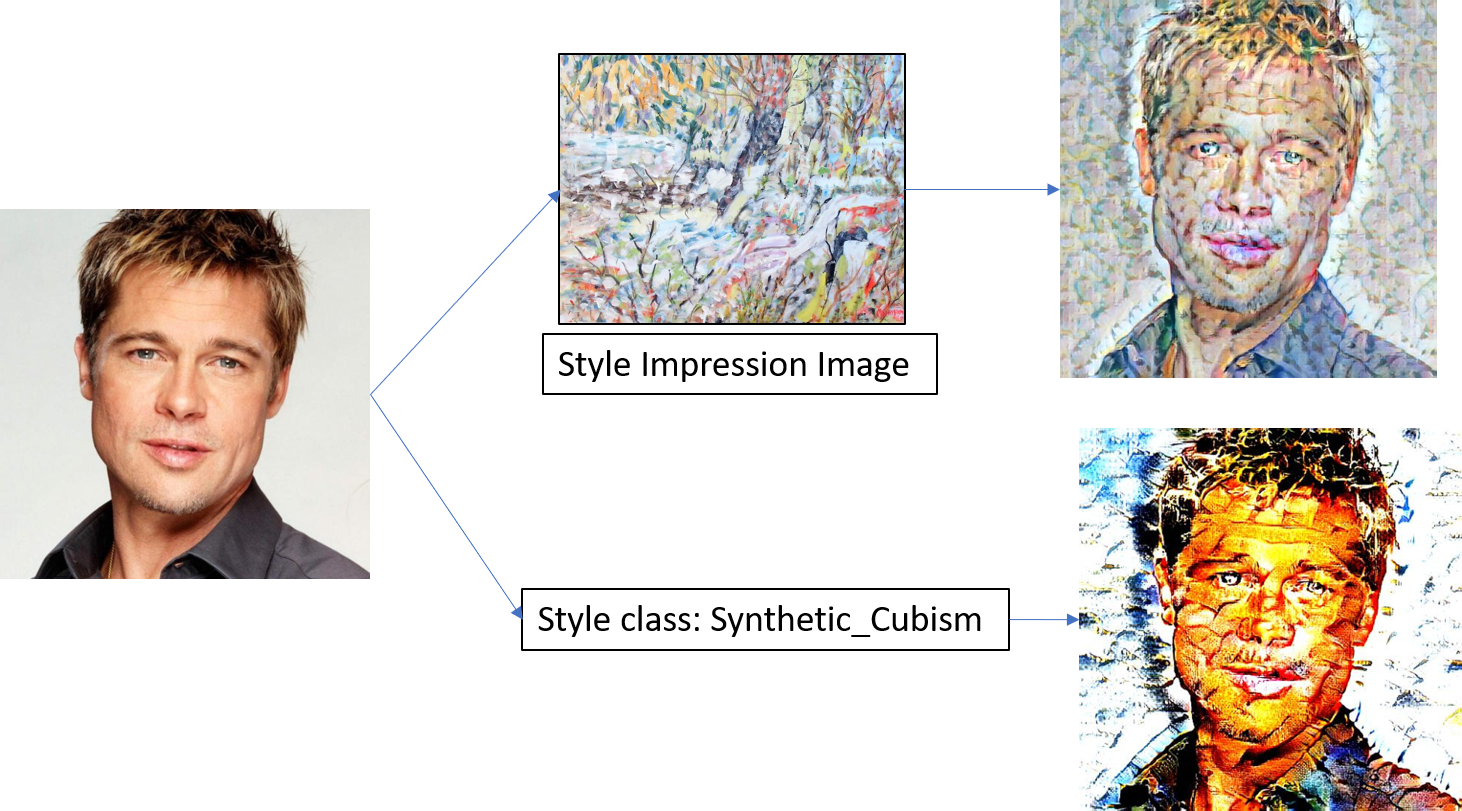}
  \caption{Image-based and Domain-based style transfer results. Note that the Domain-based transfer specified the label: Synthetic Cubism. The generative model of UST sampled AdaIN style from the label's style distribution, which comes from all the Synthetic Cubism image in the distribution generation process. Although Image-based stylize result only used one Synthetic Cubism image's style information, two results are similar in many aspects.}
  \label{fig:result1}
\end{figure}

\begin{figure*}[htbp!]
  \centering
  \includegraphics[width=0.8\textwidth]{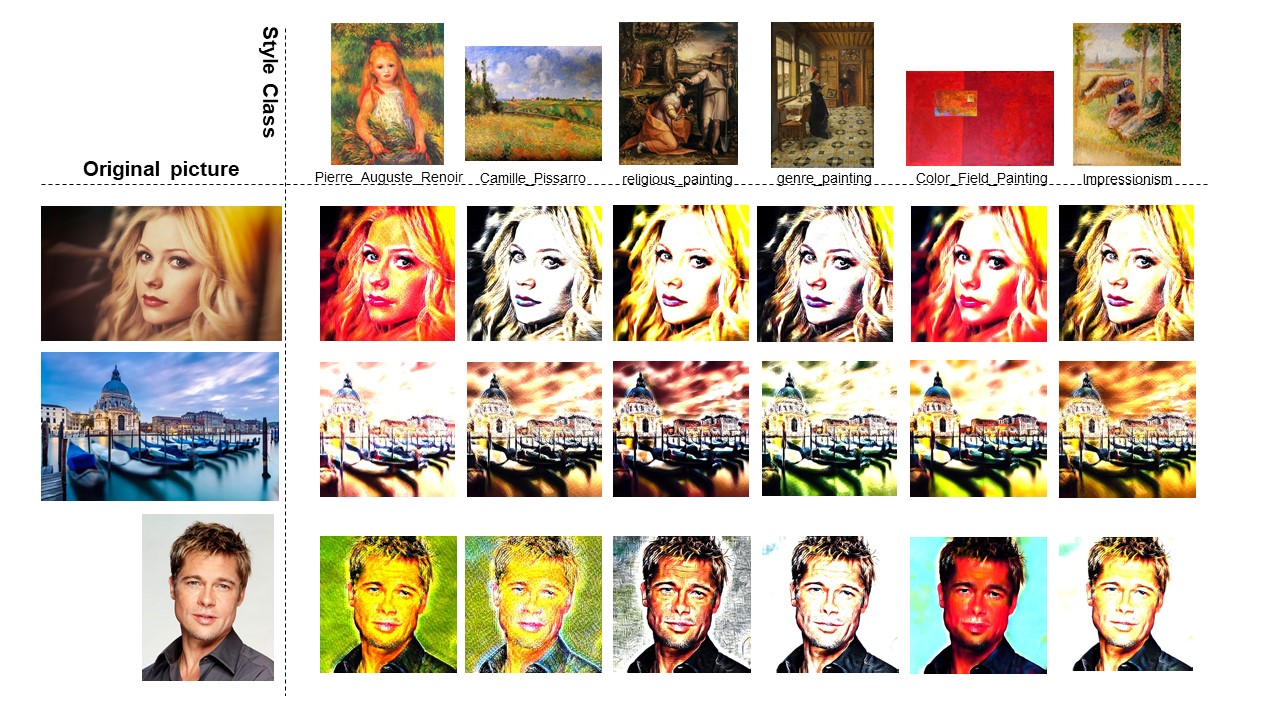}
  \caption{Domain-based transfer results. The first column is the content images, and the first row is one sample image of the class labels shown below. Note that all style information comes from the label corresponding distributions. Although it is hard to judge the performance of the results, stylized images contain many necessary style features such as colours, lines, \etc comparing with sample images. }
  \label{fig:domain_res}
\end{figure*}

In this part, we would like to display UST's style transfer results and focus on the domain-based transfer to evaluate whether multivariate Gaussian is enough to describe historical style classes' actual distribution. 
Fig.\ref{fig:result1} illustrates an example of results via two different style transfer methods of UST.
Fig.\ref{fig:domain_res} is the ablation study of Domain-based study results.

More specifically, we compare the stylized image with original class images to find the similarities and differences in low-level features. However, due to the visual check's limitations, the evaluation mainly focuses on the inclusive relationship between distributions. For example, if colour or line ambiguity appears in the generated image, the actual style distribution is beyond the Gaussian distribution description range. In contrast, if it does not, which means the actual distribution is included in Gaussian distribution.

\begin{figure*}[htbp!]
  \centering
  \begin{subfigure}{0.45\textwidth}
    \centering
    \includegraphics[width=0.9\linewidth]{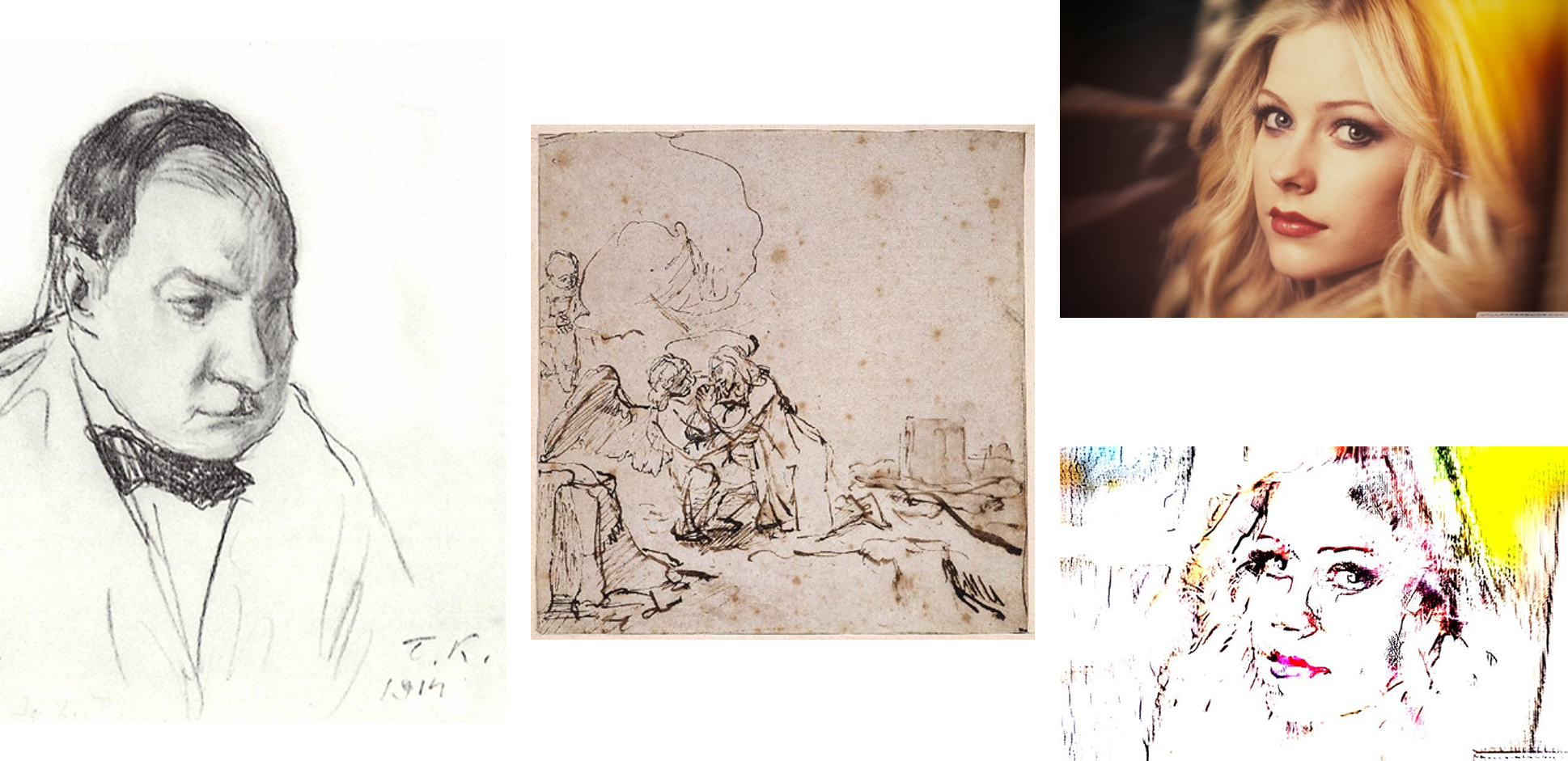}
    \caption{Gener (Sketch and Study include 2760 images)}
    \label{fig:successcase1.1}
  \end{subfigure}%
  ~ 
  \begin{subfigure}{0.45\textwidth}
    \centering
    \includegraphics[width=0.9\linewidth]{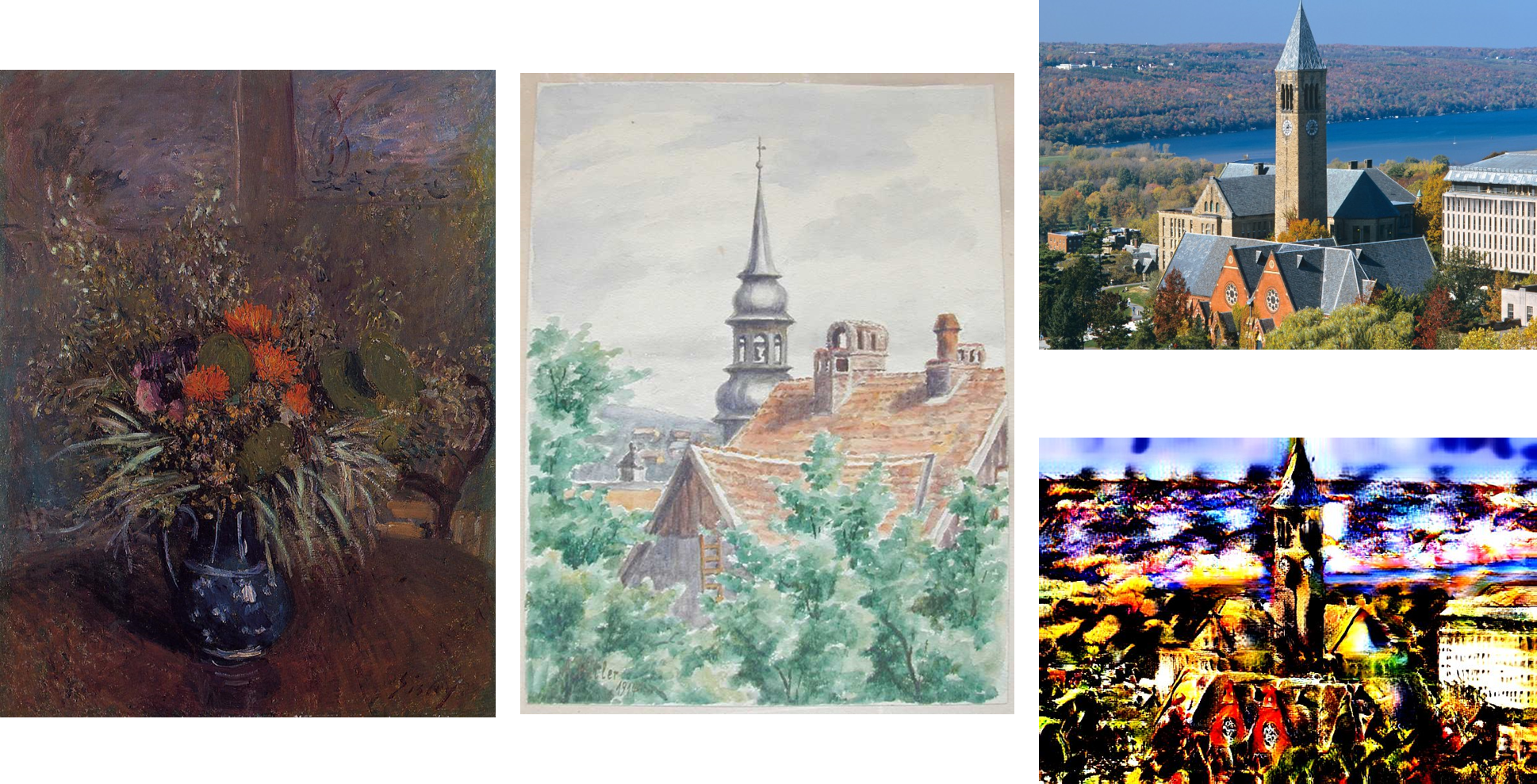}
    \caption{Style (Impression include 9136 images)}
    \label{fig:failure1.1}
  \end{subfigure}
  \caption{For each subfigure, the two example style images on the left are from WikiArt dataset. The upper right is the content image, and the bottom right image is the output stylized image from our UST model.}
  \label{fig.7}
\end{figure*}

In Fig.\ref{fig:successcase1.1}, the style, Sketch and Study, makes a brief and basic drawing of an object, which does not include rich colour. Our generative model summarizes the characteristics of line and colour feature well. However, in Fig.\ref{fig:failure1.1}, we show that Impression contains diverse colours and lines, but the output image from our generative model has colour ambiguity. 

Based on our observation of each category's output images, we find that if the colours used in the same style are more concentrated, such as Sketch and Study, Nude painting, Etc. Then it is more appropriate to use Gaussian distribution to describe the style. On the contrary, if certain styles contain various colours, such as Religious Painting, Expressionism, Etc., the Gaussian distribution may not describe this diversity well.
\subsection{Style analysis}
In the WikiArt dataset, sixty style labels are divided into three large groups: Artist, Genre and Style. Using the experiment process mentioned above, we calculate the distances between each of the two labels' Gaussian distribution and reformat them as a 60x60 distance matrix. Although the distance between different class labels might provide some surprising information, we mainly focus on evaluating the inter-distance within every three groups. Also, the Gener group images are divided by content or topic[21], which means the Gener labels here are not suitable in the discussion of styles. Therefore, in this part, only the Style and Artist groups will be discussed.

In analysing, the distributions with similar distances are supposed to be similar styles, whereas those without are regarded as irrelevant or opposite styles. To validate this hypothesis, we leveraged the historical reference and tried to prove the relation consistency in statistic results and art history.

\subsubsection{Artist}
We choose some artists as an example shown in Tab.\ref{tab:Artist distance}. It is clear that the distances between Camille Pissarro, Claude Monet and Childe Hassam are closed to each other, but not closed to Gustave Doré.
\begin{table*}[h!]
  \centering
  \resizebox{0.9\textwidth}{!}{%
  \begin{tabular}{@{}|c|c|c|c|c|c|c|@{}}
  \hline
  & Camille Pissarro & Childe Hassam & Claude Monet & Gustave Doré & Paul Cézanne & Pierre-Auguste Renoir \\ 
  \hline
  Camille Pissarro      & 0                & 362           & 283          & 2632         & 488          & 614                   \\
  Childe Hassam         & 362              & 0             & 296          & 2829         & 648          & 858                   \\
  Claude Monet          & 283              & 296           & 0            & 3229         & 634          & 603                   \\
  Gustave Doré          & 2632             & 2829          & 3229         & 0            & 2701         & 3199                  \\
  Paul Cézanne          & 488              & 648           & 634          & 2701         & 0            & 601                   \\
  Pierre-Auguste Renoir & 614              & 858           & 603          & 3199         & 601          & 0                     \\
  \hline
  \end{tabular}
  }
  \caption{Sample distances (Wasserstien metrics) between several artist painting style distributions, retrieved from the artist distance matrix.}
  \label{tab:Artist distance}
\end{table*}

According to historical records, they are all famous artists in the mid-1800s. Camille Pissarro, Claude Monet, Paul Cézanne and Pierre-Auguste Renoir are French painters. In 1859, several younger artists who prefer to paint in the more realistic style became friends while attending the school Académie Suisse, including Pissarro, Monet and Cézanne\cite{lewis2017paul}. 

Childe Hassam was an American Impressionist painter. Hassam had closed contact with French Impressionists when he took over Renoir's former studio\cite{weinberg2004childe}. Therefore, it is reasonable that the artists mentioned above have similar painting styles as they have close relationships. 

Gustave Doré was a French artist, printmaker, illustrator, comics artist, caricaturist, and sculptor who worked primarily with wood-engraving. He did not have much relationship with the other painters\cite{roosevelt1885life}.

\begin{table*}[h!]
\centering
  \resizebox{0.9\textwidth}{!}{
  \begin{tabular}{@{}|c|c|c|c|c|c|c|@{}}
  \hline
  & Colour Field Painting & Early Renaissance & Minimalism & Realism & Romanticism & Ukiyo-e \\  
  \hline
  Colour Field Painting & 0                    & 3337              & 892        & 896     & 1137        & 5456    \\
  Early Renaissance      & 3337                 & 0                 & 2973       & 1569    & 1095        & 1173    \\
  Minimalism             & 892                  & 3973              & 0          & 1508    & 1460        & 4766    \\
  Realism                & 896                  & 1569              & 1508       & 0       & 101         & 3511    \\
  Romanticism            & 1137                 & 1095              & 1460       & 101     & 0           & 2877    \\
  Ukiyo-e                & 5456                 & 1173              & 4766       & 3511    & 2877        & 0       \\ 
  \hline
  \end{tabular}
  }
  \caption{Sample distances (Wasserstien metrics) between several historical style distributions, retrieved from the style distance matrix.}
  \label{tab:style distance}
\end{table*}
  
\subsubsection{Style}

\begin{figure}[htbp!]

  \centering
  \begin{subfigure}{0.3\textwidth}
    \centering
    \includegraphics[width=0.9\linewidth]{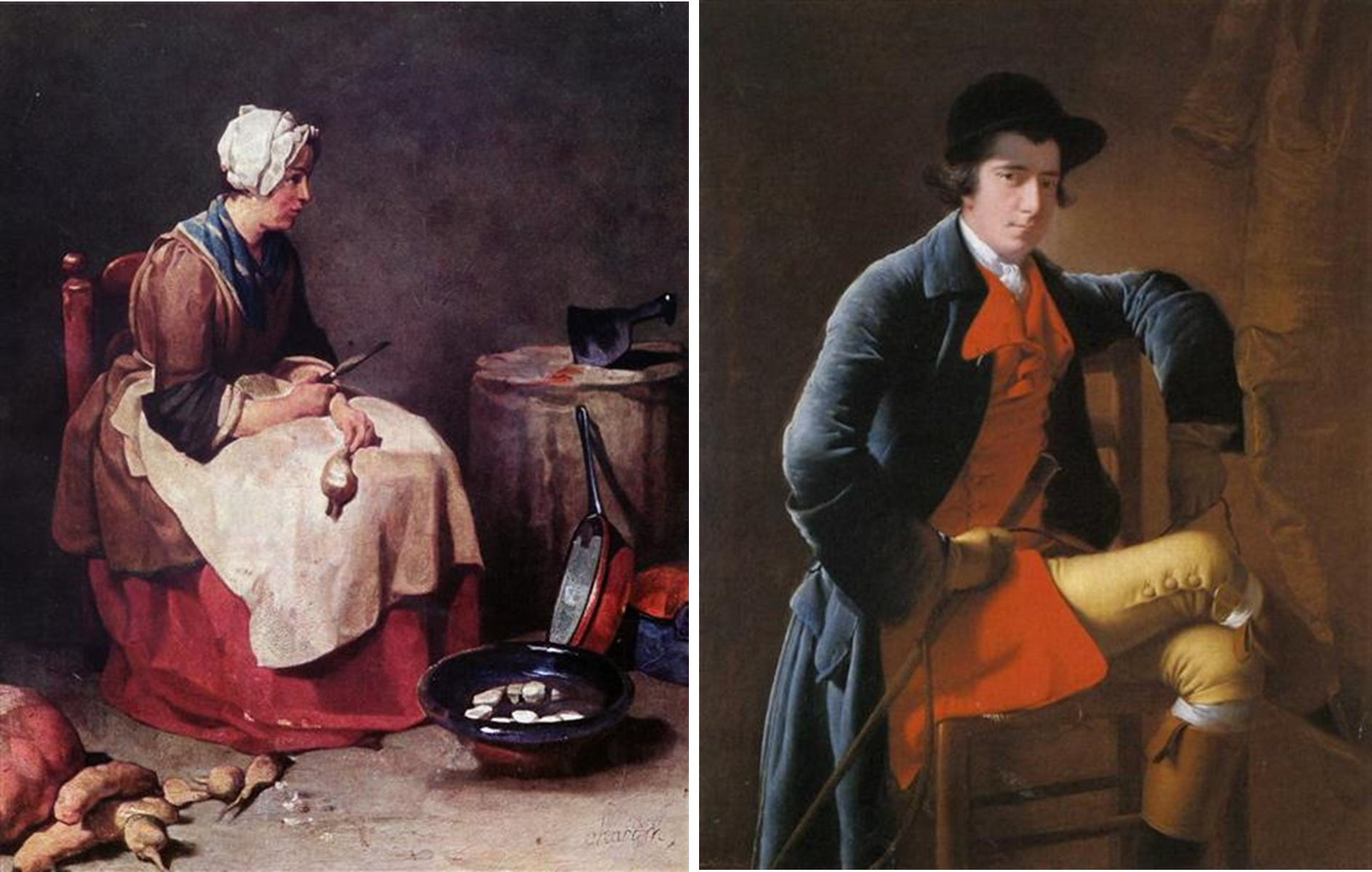}
    \caption{Realism and Romanticism have similar texture}
    \label{fig:Style1}
  \end{subfigure}%
 
  \begin{subfigure}{0.1\textwidth}
    \centering
    \includegraphics[width=0.9\linewidth]{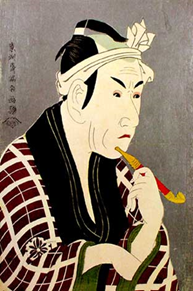}
    \caption{Ukiyo-e}
    \label{fig:Style2}
  \end{subfigure}
  ~
  \begin{subfigure}{0.15\textwidth}
   \centering
   \includegraphics[width=0.9\linewidth]{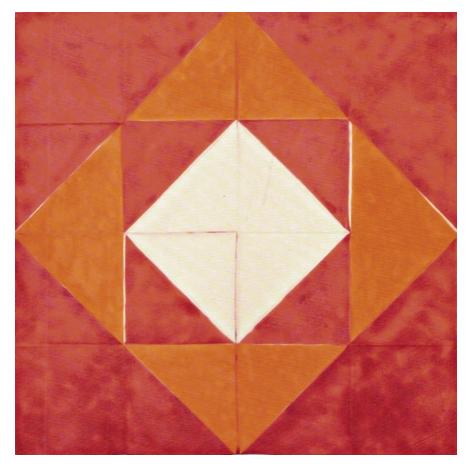}
   \caption{Colour Field painting and minimalism}
   \label{fig:Style3}
 \end{subfigure}
  \caption{Style example images}
  \label{fig:Styletotal}
\end{figure}

We choose some styles as an example shown in Tab.\ref{tab:style distance}, the distance between Realism and Romanticism distributions is only 101, which is the smallest. Colour Field Painting and Minimalism is also close to each other, but Ukiyo-e is far from the other styles. The human-sense style label here is not limited to the description of low-level features. Besides, it is a combination of history and cultural backgrounds\cite{fernie1995art}. Sample images in each style label mentioned above show in Fig.\ref{fig:Styletotal}.

It is easy to find evidence that Colour Field Painting and Minimalism are both originated in America in the mid-1900s \cite{wilkin2007color}. Note that the example image Fig.\ref{fig:Style3} is below to both of the styles. However, Realism was considered a polar opposite to Romanticism in art theory \cite{turgenev1965fathers}, which is ostensibly contrary to our results. To explain this, Realism and Romanticism originated in Europe in the 18th-19th century. The central conflict between them is the spirit and stories in the paintings but not the manifestations. Therefore, we conclude that they influenced each other in some low-level features, such as painting techniques. Ukiyo-e is a Japanese style that is not relevant to most of the western-origin styles.\cite{harris2012ukiyo}, which is also consistent with our results.

\subsection{Experiment summary}
To summarize the Experiment Results part, we use results from Unified Style Transfer and Statistical Style Analysis to validate two hypotheses \ie the ``if and only if'' relationship between similar styles and close distributions, from equation.\ref{equation:1} and whether the Gaussian distribution is good enough to describe actual distribution in AdaIN vector space. The results show that the statistical distances are consistent with the human-sense of style relations, which is a strong supporter of our first hypothesis' forward direction. The results also show that Gaussian distribution has limitations in representing actual distribution in AdaIN style space. It could cover necessary low-level features of a specific style class but fail to handle the balance of colour or the high-level features.

These two conclusions help us evaluate our model \ie UST's performance and reveals the limitations of using Gaussian distribution. Also, it proves that the Statistical Style Analysis method is robust in validating the models/distributions feasibility. 

\section{Discussion and Conclusion}
\label{sec:conclusion}

Our study develops the Unified Style Transfer model that can conduct both Image-based and Domain-based style transfer simultaneously. Besides, we propose a new philosophy to evaluate style transfer models called Statistical Style Analysis. We raise two hypotheses on style representations and prove them by experiments.

Our work could shed light on what the style of an image truly means during style transfer. We can try more distribution models and style representations on some more comprehensive artwork datasets in the future. Using unsupervised or semi-supervised learning method can be one of our future options.
\clearpage

{\small
\bibliographystyle{ieee_fullname}
\bibliography{egbib}
}

\end{document}